\documentclass{ifacconf}

\usepackage{graphicx}      
\usepackage{natbib}        
\usepackage{comment}
\usepackage{amsfonts}
\usepackage{amsmath}
\usepackage{amssymb}
\usepackage{array}
\usepackage{cancel}
\usepackage{subcaption}
\usepackage{color}
\usepackage{epsfig}

\usepackage{tikz} 
\usepackage{pgfplots}
\pgfplotsset{compat=1.17}

\begin{document}
\begin{frontmatter}

\title{Characterizing Gaussian Mixture of Motion Modes for Skid-Steer Vehicle State Estimation\thanksref{footnoteinfo}} 



\thanks[footnoteinfo]{DISTRIBUTION A. Approved for public release; distribution unlimited.
OPSEC9637}

\author[First]{Ameya Salvi} 
\author[Second]{Mark Brudnak}
\author[Second]{Jonathon Smereka} 
\author[First]{Matthias Schmid}
\author[First]{Venkat Krovi}

\address[First]{Department of Automotive Engineering,Clemson University, SC, 29634, USA}
\address[Second]{U.S. Army DEVCOM Ground Vehicle Systems Center, Warren, MI 48092, USA }

\begin{abstract}                

The skidding and slipping motion of skid-steered wheel mobile robots (SSWMRs) is highly influenced by the complex nature of tire-terrain interactions. The lack of reliable terrain friction models cascade into unreliable motion models, especially the reduced ordered variants used for state estimation and robot control. Ensemble modeling is an emerging research direction where the overall motion model is broken down into a family of local models to distribute the performance and resource requirement and provide a fast real-time prediction. To this end, a Gaussian Mixture Model (GMM) based modeling approach for identification of model clusters is adopted and implemented within an Interactive Multiple Model (IMM) based state estimation framework. The methodology is adopted and implemented for estimating angular velocity for a mid scale skid-steered wheel mobile robot platform.

\end{abstract}

\begin{keyword}
Clustering, Mixture Models, State Estimation, Skid-steered robots
\end{keyword}

\end{frontmatter}

\section{Introduction}~\label{sec:Introduction}
The rugged nature of skid-steered wheel mobile robots (SSWMRs) enable them to be instrumental in strenuous environments such as mining, construction and agriculture. Operationalizing fully autonomous SSWMRs is thus critical for alleviating personnel challenges commonly witnessed in such challenging scenarios. Unfortunately, minimal human supervision entails detailed and systematic investigation of the autonomy characteristics of SSWMRs which vary significantly across the robot's size, scale and operation regimes. An essential aspect of such an analysis is the determination of analytical motion models of the robot that are necessary for autonomy modules such as state estimation, localization and model based controls. 

Motion models for ground vehicles have been investigated extensively in the context of Ackermann steered vehicles and wheeled mobile robots, both in the aspects of models for motion analysis and reduced ordered models for controls and estimation~[\cite{Jazar2019,SiegwartAMR}]. Unfortunately, SSWMRs posses a unique challenge due to steering free nature of the robot that relies on skidding for executing motion maneuvers. The robot's motion is thus dictated by the friction dominant tire-terrain interactions, capturing which are critical for accurate identification of the robot motion models. 

\begin{figure}[t]
    \centering
    \includegraphics[width=0.99\linewidth]{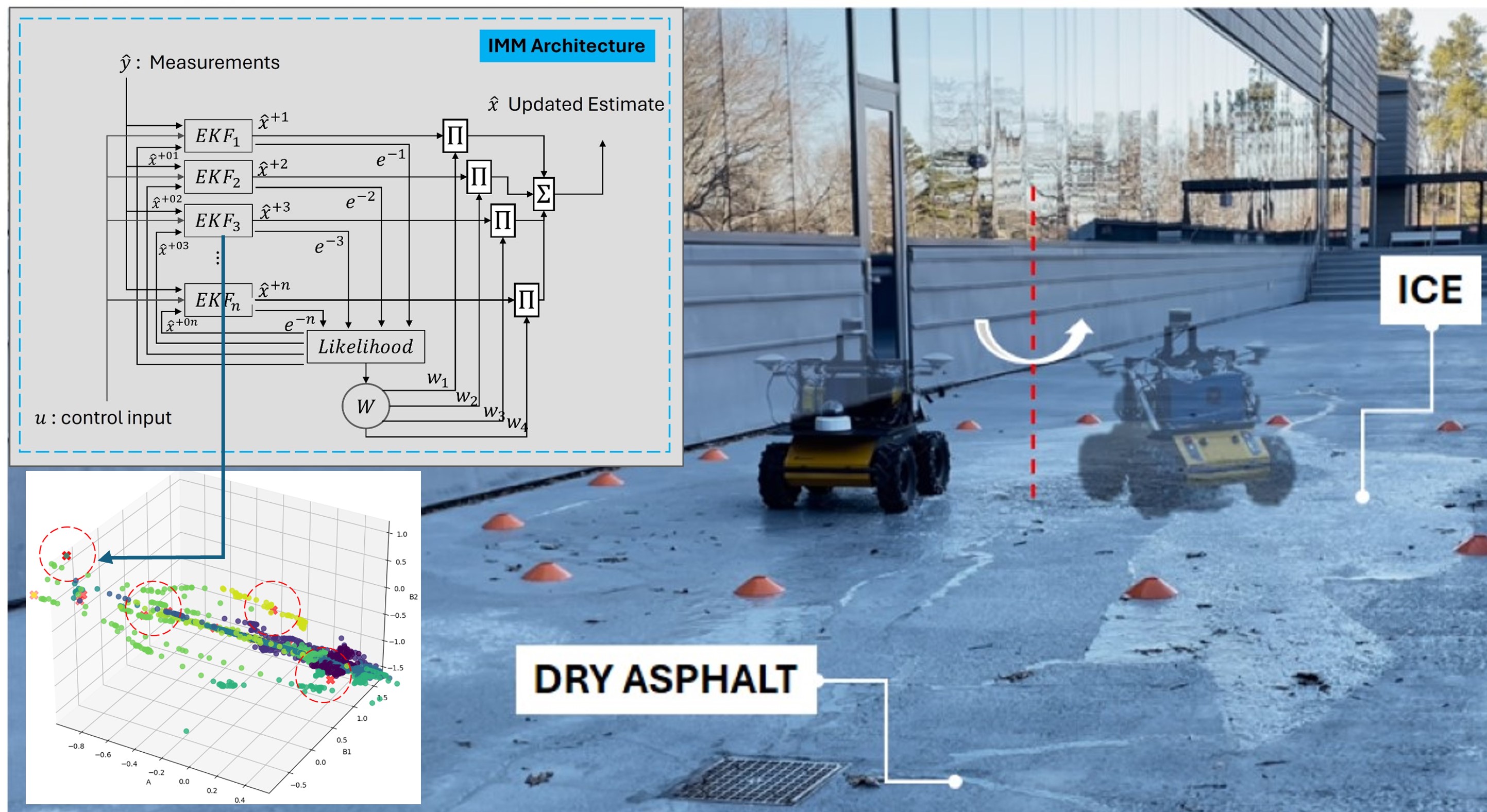}
    \caption{Overview of the interactive multiple model (IMM) based state estimation framework utilizing the motion models represented by the mixture of Gaussian to compensated for the aggravated skidding on ice.}
    \label{fig:GMM-IMMOverview}
\end{figure}

Out of the body of work investigating SSWMR motion models discussed in the section~\ref{sec:Literature}, a large portion now focus on parameter identification or calibration based models that allow for \textit{light-weight} models for real-time state-estimation and controls. While the approach is sound, it is limited by the engineer's domain expertise in identifying the model's basis, which almost always is inadequate in providing a proper fit across the entire calibration dataset. To this end, a Gaussian mixture model (GMM) based model clustering approach has been proposed in this work. In particular :

\begin{itemize}
    \item A mixture of Gaussian in used to represent a family of linear motion models instead of a single model.
    \item The models are utilized with interactive multiple model based state-estimation framework and extensively investigated for state estimation performance against a standard single model based Kalman filter.
\end{itemize}

\section{Related Literature}~\label{sec:Literature}
Kalman filters (KF) follow a rich history as an optimal unbiased state estimator and have been studied extensively over the years in academic and real-time applications. KF fuse the predictions from a linear prediction model with the measurements to estimate the real-time value of the state in concern. In the context of state estimation for robotics, KF, and its non-linear variants such as the extended Kalman filter (EKF), unscented Kalman filter (UKF), among many others have been investigated for robot pose and velocity estimation~[\cite{crassidis2004optimal,thrun2002probabilistic}]. While the non-linear variants alleviated the need for a linear prediction model, they introduce significant level of complexity for tasks such as equivalent linear approximations, filter tuning for highly non-linear systems, and, can introduce real-time computational complexity for resource constrained deployments~[\cite{simon2006optimal}]. These challenges can be aggravated in the context of SSWMRs which do not have a reliable motion model (linear or non-linear) thus making the non-linear state estimation even more challenging. Thus, improving the efficacy of KFs, especially in the context hard to model SSWMRs can contribute significantly for their real-time state estimation.

One peculiar challenge brought in by linear prediction models is that their performance is limited over the entire operation domain, especially for highly non-linear systems. A typical approach adopted to mitigate this issue is utilization of multiple linear models defined over different operational regimes and then bringing them together with mixing of state estimates using interacting multiple models state estimation(IMMs)~[\cite{RAMAN2022369,Gill2019,salvi2024IMM}]. While the multiple-model based state estimation has mostly been investigated for capturing the environment driven model changes, its applicability to overcome system's non-linearity has yet to be investigated. The SSWMR skidding on ice presented in this work provides a unique combination of environmental effects and system specific non-linearities thus providing a suitable scenario to investigate the IMM base state estimation. 

SSWMR models have been extensively investigated in the context of reduced ordered kinematic formulations~[\cite{Mandow2007a, Wang2015, Rabiee2019a, Ordonez2017}], both as linear and non-linear approximations for the estimating robot motion mechanics, while some investigating the validity of these models for extreme conditions~[\cite{Baril2020}]. The unpredictable nature of skidding outlined in all approaches lead to the utilization of some form of data-fitting approaches to tune the proposed models. Such a tuning and calibration requirement introduces issues associated with quality of data collection and its pre-processing. When put in the context of identifying several linear models, the necessity to accurately define the operation regions is introduced. Thus, clustering the relevant data samples for identifying an accurate locally linear model is a critical challenge that needs to be addressed.

Supervised and un-supervised data clustering approaches have gained popularity within the machine learning community in recent times~\cite{alloghani2020clustering,sindhu2020clustering}. Compared to the supervised clustering methods, the un-supervised clustering approaches such as k-means clustering, Gaussian mixture models and principle component analysis (PCA) can be helpful for machine learning based automated identification of data-clusters thus eliminating the human bias in the framework. While data-driven machine learning approaches such as Gaussian process regression and physics informed machine learning have been investigated for model identification, the utilization of un-supervised clustering for aggregating linear models is yet to be investigated. 

To this end, a combination of GMM based linear models with IMM estimation is proposed in this work. In particular, the influence of number of components of the GMM clustering is investigated for state-estimation performance. 

\section{Problem Formulation}~\label{sec:ProblemFormulation}
The state estimation problem selected for illustrating the proposed framework is for estimating the angular velocity for a Clearpath skid-steer husky robot. The robot is integrated with a 9 axis IMU that provides measurements for the angular velocity and linear accelerations. The robot is operated on an icy surface that creates intermittent aggravated skidding scenarios where the simplified linear representation for the SSWMRs is insufficient for state estimation. Due to absence of linear velocity for measurements or model identification, angular velocity is utilized as the sole state for estimation.

\subsection{Discrete time motion models}~\label{sec:MotionModels}
The discrete time representation of a continuous time model is the standard formulation utilized in most of recursive state estimation framework~[\cite{crassidis2004optimal}]. Such a formulation for a linear time system can be realized by applying zero order hold on the control input to achieve the following formulation : 

\begin{equation}
\label{eq:DiscreteModel}
    \begin{aligned}
    &\mathbf{x}_{k+1} = \mathbf{x}_{k} + \mathbf{A}_{d}\mathbf{x}_{k} + \mathbf{B}_{d}\mathbf{u}_{k} + \mathbf{w}_{k} \\
    &\mathbf{y}_{k+1} = \mathbf{C}_{d}\mathbf{x}_{k} + \mathbf{D}_{d}\mathbf{u}_{k} + \mathbf{v}_{k}
    \\
     \end{aligned}
\end{equation}

The state estimation for the angular velocity, $\omega$, is represented by state $\mathbf{x} \in \mathbf{R}^1$, and, $\mathbf{A}_d, \mathbf{B}_d, \mathbf{C}_d$ and $\mathbf{D}_d$ are linear matrices representing the discrete time dynamical system. These matrices are identified using linear least squares approximation over the robot trajectory dataset collected for control inputs $\mathbf{u} = [\dot{\phi}_l \quad \dot{\phi}_r]^T$ representing the left and right wheel input velocities. $\mathbf{w}$ and $\mathbf{v}$ represent the process and measurement noises that are tuned heuristically for the implementation. The discrete time model represents state transition for any timestep $k$ to the next timestep $k+1$. For this work, the measurement model $\mathbf{C}_d$ is an one dimensional identity matrix and $\mathbf{D}_d$ is $1 \times 2$ null matrix.

\subsubsection{Dataset and model fitting}~\label{sec:Dataset}

\begin{figure}[t]
    \centering
    \begin{subfigure}[b]{0.95\linewidth}
        \centering
        \includegraphics[width=\linewidth, height = 6cm]{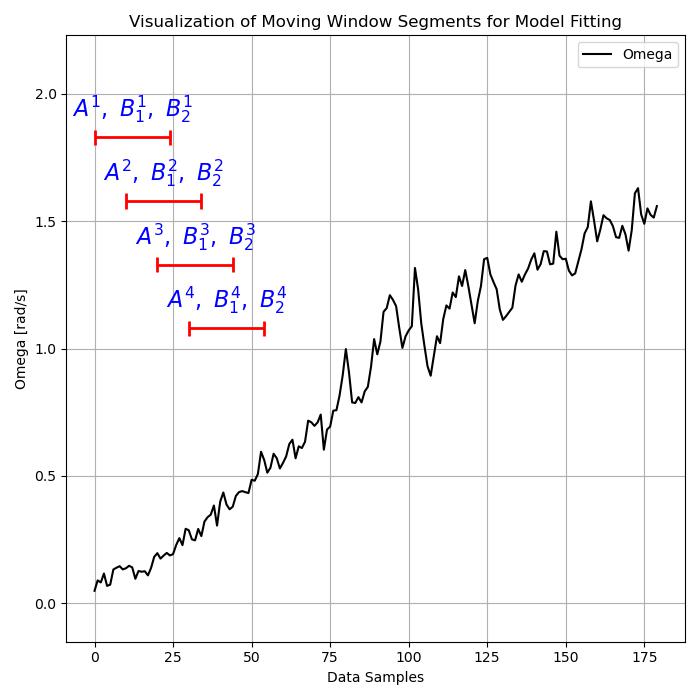}
        \caption{Sliding windows of a sequence of 25 samples of angular velocity used to fit the matrices $A$ and $B$. The methodology extends over the entire dataset of 9 such trajectories.}
        \label{fig:DataSetWindow}
    \end{subfigure}
    
    \vspace{0.5cm} 

    \begin{subfigure}[b]{1\linewidth}
        \centering
        \includegraphics[width=\linewidth]{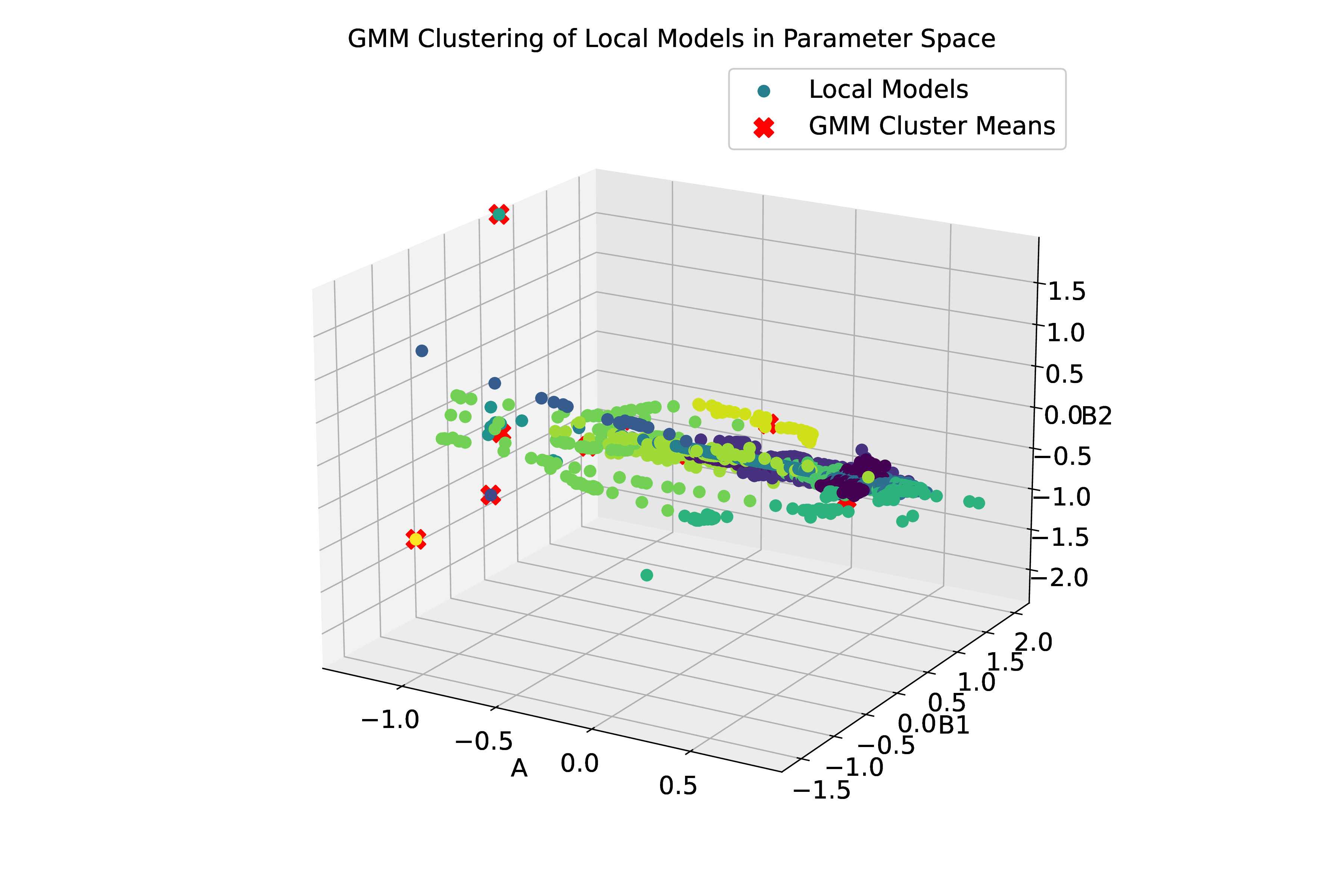}
        \caption{Each sequence of trajectory from figure~\ref{fig:DataSetWindow} is used to realize a single linear model. The figure illustrates collections of such linear models with each sample point in the figure as one model. 3000 models are realized over the entire collection of dataset.}
        \label{fig:ClusterSpace}
    \end{subfigure}
    
    \caption{Representing linear models in the parameterized with the fitting matrices $A$ and $B$}
    \label{fig:LinearModelRepresentation}
\end{figure}

For a given vehicle trajectory dataset  of $N_D$ samples:

\begin{equation}
    \tau = {[\mathbf{x}_k, \mathbf{u}_{k},\mathbf{x}_{k+1}]}_{0}^{N_D}
\end{equation}

The linear models $\mathbf{A}_d$ and $\mathbf{B}_d$ ($\mathbf{B}_d = [B_1 \quad B_2]$), can be identified as : 

\begin{align}
X^+ = \begin{bmatrix} A & B \end{bmatrix} \begin{bmatrix} X \\ U \end{bmatrix}
\end{align}
where
\begin{align}
X &= \begin{bmatrix} x_0 & x_1 & \cdots & x_{N-1} \end{bmatrix} \in \mathbb{R}^{1 \times N} \\
U &= \begin{bmatrix} \phi_L \\ \phi_R \end{bmatrix} \in \mathbb{R}^{2 \times N} \\
X^+ &= \begin{bmatrix} x_1 & x_2 & \cdots & x_N \end{bmatrix} \in \mathbb{R}^{1 \times N}
\end{align}

Fitting global linear models $A_g$ and $B_g$ over the entire trajectory dataset while convenient is often inadequate. A common solution proposed in literature are fitting locally linear models over subsets of the trajectory. For prediction and control, these models are strategically chosen depending on the operating conditions.

\subsection{Gaussian Mixture Models}~\label{sec:GMM}
Unfortunately, it can be significantly challenging to identify how to split the dataset for identifying these multiple models. Majority of the methods in the literature rely on engineering approximations to identify varied operating conditions to identify dataset splits. To alleviate this challenge, an incrementally sliding window approach is utilized to define the trajectory's sample set. A local model is then fit for each of the defined window, subsequently allowing to realized several locally linear models. Figure~\ref{fig:DataSetWindow} illustrates the windowing method for setting up the local trajectory sequence. Figure~\ref{fig:ClusterSpace} defines the locally linear models for each of the data sequence. Each data point $s^n = [A^n, B_{1}^n, B_{2}^n]$ is representative of one locally linear model. For this work, a sequence of $25$ data points incrementing the window with one data point is used to realize $3000$ locally linear models.

The distribution of models illustrated in figure~\ref{fig:ClusterSpace} indicates the shear number of local models which can make it computationally difficult to utilize all in real-time. Interestingly, the distribution also indicates the local concentration of the models which allows to utilize clustering approaches to reduce the number of models thus making it feasible to utilize the multiple models for estimation.

\begin{figure*}
    \centering
    \includegraphics[width=1\linewidth]{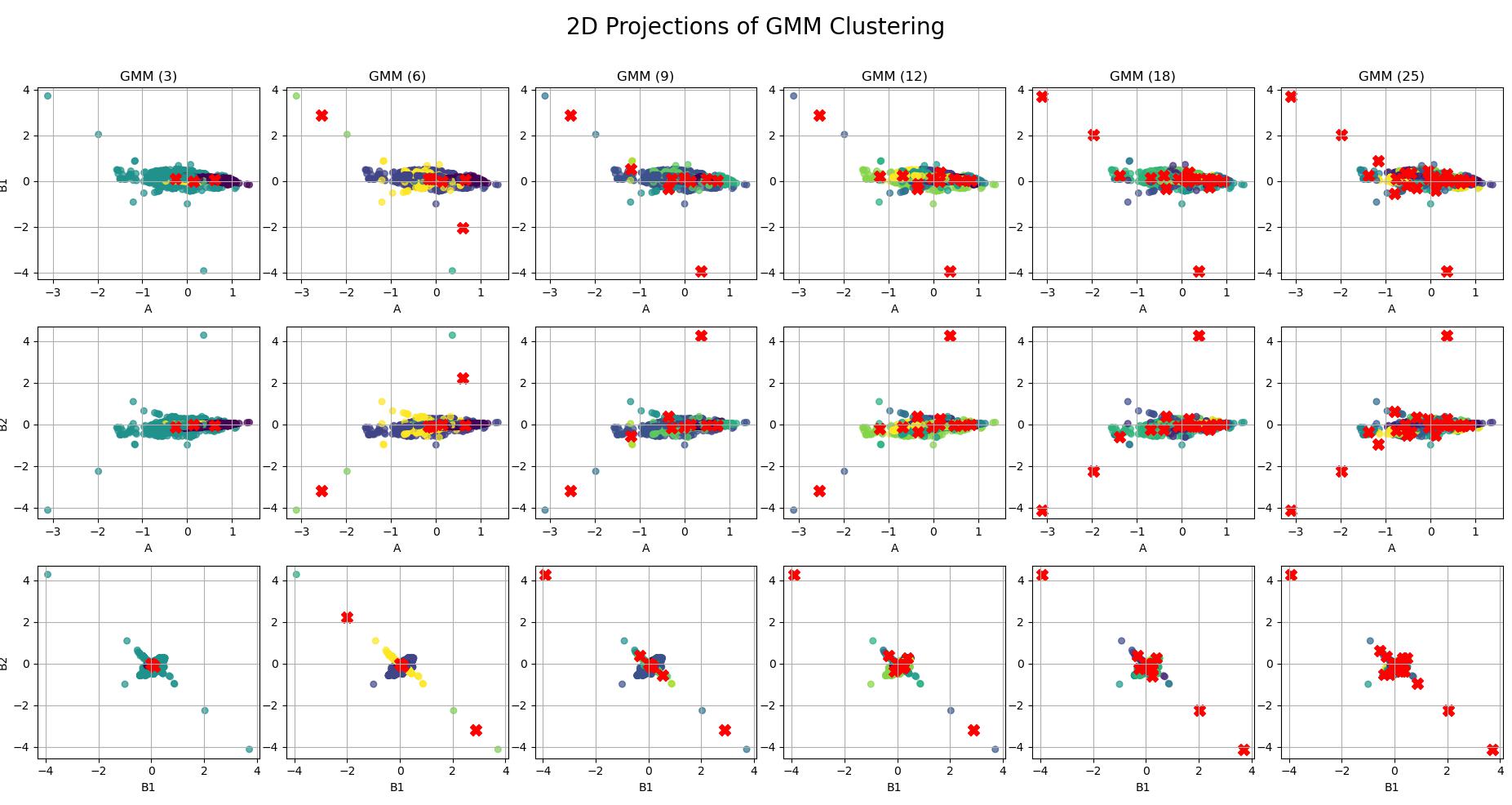}
    \caption{2D Projections of GMM based unsupervised clustering for the the model parameters $A$, $B_1$ and $B_2$. (Left to right) Clustering results when increasing the number of Gaussian components}
    \label{fig:GMMFit2D}
\end{figure*}

Mixture models adopt an expectation maximization paradigm (EM) to assign likelihood of belonging to a specific cluster to each data point. For a given unlabeled dataset of size $N$, $S = [s_1, s_2, ... s_{N}]$, and pre-defined number of components $M$, The responsibilities for any $n^{th}$ data sample for $m^{th}$ component ($m \in [1,M]$) at any iteration step $t$ is given as:
\begin{align}~\label{eq:responsibilities}
\gamma_m^{(t)}(s_n) = \frac{ \pi_m^{(t)} \cdot \mathcal{N}(s_n \mid \mu_m^{(t)}, \Sigma_m^{(t)}) }{ \sum_{j=1}^{M} \pi_j^{(t)} \cdot \mathcal{N}(s_n \mid \mu_j^{(t)}, \Sigma_j^{(t)}) }
\end{align}

where, $\pi_m$ is the $m^{th}$ component weight, $\mu_m$ is the $m^{th}$ component mean and $\Sigma_m$ is the $m^{th}$ component covariance. $\mathcal{N}$ follows standard notation and representation for normal distribution. In the initialization step, all the weights are initialized as same ($\pi_j = 1/M$) and the means and covariances, $\mu_j$ and $\Sigma_j$, initialized randomly ($\forall j \in [1,M]$).

At the next step $t+1$, the responsibility count $r_m$ (indicating the number of samples from the dataset belonging to the $m^{th}$ component), the component weights $\pi$, means $\mu$ and covariances $\Sigma$ are updated as :


\begin{align}
r_m^{(t+1)} &= \sum_{n=1}^{N} \gamma_m^{(t)}(s_n) \\
\pi_m^{(t+1)} &= \frac{r_m^{(t+1)}}{N} \\
\mu_m^{(t+1)} &= \frac{1}{r_k^{(t+1)}} \sum_{n=1}^{N} \gamma_m^{(t)}(s_n) \cdot s_n
\end{align}
\begin{equation}
\begin{aligned}
\Sigma_m^{(t+1)} &= \frac{1}{r_m^{(t+1)}} \sum_{n=1}^{N} \gamma_m^{(t)}(s_n) \cdot \\
                 &(s_n - \mu_m^{(t+1)})(s_n - \mu_m^{(t+1)})^T
\end{aligned}
\end{equation}
The updated weights, means and covariances are used to re-calculate the responsibilities in equation~\ref{eq:responsibilities} and then again updated recursively until convergence. 

Figure~\ref{fig:GMMFit2D} represent the model space projected as 2D plots of $A$ vs $B_1$, $A$ vs $B_2$, and, $B_1$ vs $B_2$. The figure represents the locations of cluster means as the number of components increase. Each data point is three dimensional, and the covariance for all fittings is assumed to be diagonal for ease of computation. The choice and granularity of the number of components (3 to 25) is a design choice and can be a subject of independent study. 

\subsection{Interactive Multiple Model Estimation}\label{sec:IMM}

Multiple model estimation frameworks such as interactive multiple model-estimation (IMM) and multiple model adaptive estimation (MMAE) implement a bank of filters with each providing a prediction dependent on the model adopted by that filter for prediction. The predictions are fused with the real-time measurements and then mixed depending on the model weights to provide a reliable state estimation. The dynamically varying weights for the filter (especially in the IMM framework), allow to keep switching between the models to provide much accurate state estimation. The key steps of the IMM filter are outlined below:

\subsubsection{\textbf{Mixed state and covariance estimated}}
\begin{align}~\label{eq:filter_initializations}
\bar{\mathbf{x}}_k^{(m)} &= \sum_{i=1}^{M} \mu_{k-1}^{(i)} Tr_{ij} \, \mathbf{x}_{k-1}^{(i)}
\end{align}
\begin{equation}
\begin{aligned}
\bar{\mathbf{P}}_k^{(i)} &= \sum_{i=1}^{M} \mu_{k-1}^{(i)} Tr_{ij} \\
&\left( \mathbf{P}_{k-1}^{(i)} + (\mathbf{x}_{k-1}^{(i)} - \bar{\mathbf{x}}_k^{(i)})(\mathbf{x}_{k-1}^{(i)} - \bar{\mathbf{x}}_k^{(i)})^T \right)
\end{aligned}
\end{equation}

At every step $k$, for every $i^{th}$ model, the state estimates $\bar{\mathbf{x}}_k^{(i)}$ and covariance $\bar{\mathbf{P}}_k^{(i)}$ are prepared based on the model weight $\mu$ and the state transition probability $Tr$ for estimating the state priors. The state transition probability $Tr_{ij}$ defines the probability of transitioning from model $i$ to model $j$ which is typically pre-defined. The model weights are initialized equally during the first steps and gets updated dynamically through the process.

\subsubsection{\textbf{Filter update for each model}}
\begin{align}
\hat{\mathbf{x}}_k^{(i)} &= \text{KalmanFilter}(\bar{\mathbf{x}}_k^{(i)}, \bar{\mathbf{P}}_k^{(i)}, \mathbf{z}_k, [ \mathbf{A} \quad \mathbf{B}]^i)
\end{align}

Based on the earlier step, the priors are calculated utilizing the standard Kalman filtering approach~\cite{Crassidis} for each models. The state priors are calculated utilizing the distinct models $[\mathbf{A} \quad \mathbf{B}]^{i}$ identified using the GMM clustering and then updated using same measurements $\mathbf{z}$.


\begin{figure*}[t!]
    \centering
    \includegraphics[width=1\linewidth]{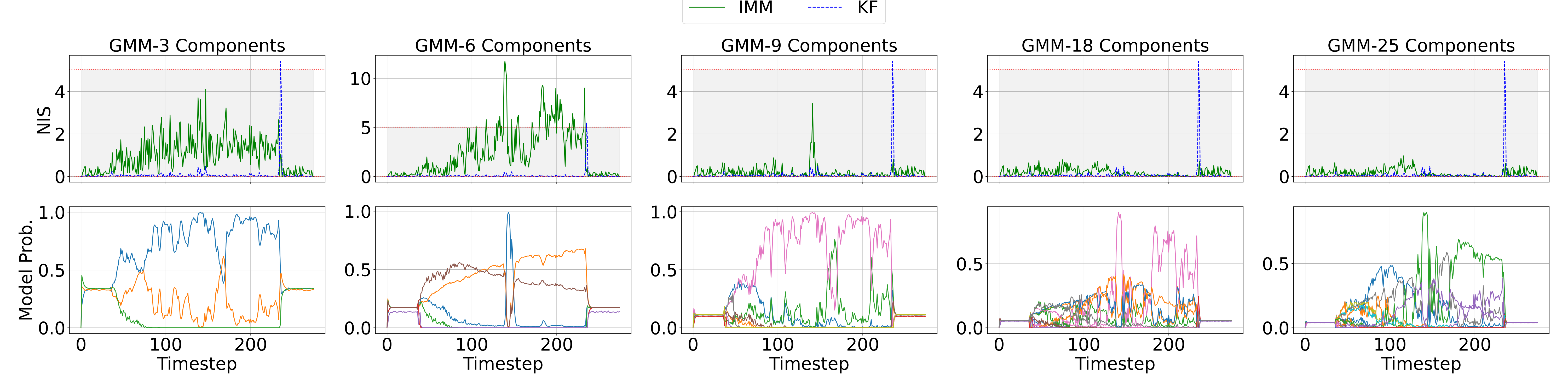}
    \caption{Comparing the NIS statistics when estimating of angular velocity using (a)single model Kalman filter, and (b) IMM framework utilizing models clustered using mixture of Gaussian. (Left to Right) Visualizing the impact of increasing the number of components of the Gaussian mixture. (Second Row) Dynamically updating model weights illustrating the changing model requirement for improved state estimation.}
    \label{fig:enter-label}
\end{figure*}


\subsubsection{\textbf{Likelihood and Model probabilities}}
\begin{align}
\mathcal{L}_k^{(i)} &= \frac{1}{\sqrt{2\pi \mathbf{S}_k^{(i)}}} \exp\left( -\frac{(\mathbf{y}_k^{(i)})^2}{2 \mathbf{S}_k^{(i)}} \right)
\end{align}
where, \[ \mathbf{y}_k = \hat{\mathbf{x}}_k^i - \mathbf{z}_k \]
\begin{align}
\mathbf{w}_k^{(i)} &= \frac{\mathcal{L}_k^{(i)} \, \mathbf{w}_{k-1}^{(i)}}{\sum_{j=1}^{M} \mathcal{L}_k^{(j)} \, \mathbf{w}_{k-1}^{(j)}}
\end{align}

Based on the updated estimates from the different filters, the likelihood $\mathcal{L}$ can be calculated for each model. Physically, the likelihood values for each model $\mathcal{L}_k$ tries to estimate how close the value of the filter estimate is to the measurement. Thus, the likelihood values $\mathcal{L}_k$ are further used for updating the model weight $\mathbf{w}^i_k$.

\subsubsection{\textbf{Combined estimated}}
\begin{align}
\hat{\mathbf{x}}_k &= \sum_{i=1}^{M} \mathbf{w}_k^{(i)} \, \mathbf{x}_k^{(i)}\\
\hat{\mathbf{P}}_k &= \sum_{i=1}^{M} \mathbf{w}_k^{(i)} \left( \mathbf{P}_k^{(i)} + (\mathbf{x}_k^{(i)} - \hat{\mathbf{x}}_k)(\mathbf{x}_k^{(i)} - \hat{\mathbf{x}}_k)^T \right)
\end{align}

Finally, the combined estimated state and covariance estimate is calculated by weighted addition of the individual filter estimates which is the recursively utilized in the next filter update steps by multiplying with the transition probabilities outlined in equation~\ref{eq:filter_initializations}.

\section{Results}~\label{sec:Results}
The GMM-IMM combined framework has been implemented for the state-estimation in the problem setup described in section~\ref{sec:ProblemFormulation}. Typically the validation of state-estimation performance can by done in the presence of ground truth information that avaliable from other accurate measurement sources such as real-time kinetic (RTK) - GPS measurements or optical tracking systems. In the absence of ground-truth, filter validation relies on the measurement data alone. One reliable approach to investigate filter performance is the characterization of the normalized innovation squared (NIS) measurements.

\begin{figure}
    \centering
    \includegraphics[width=1\linewidth]{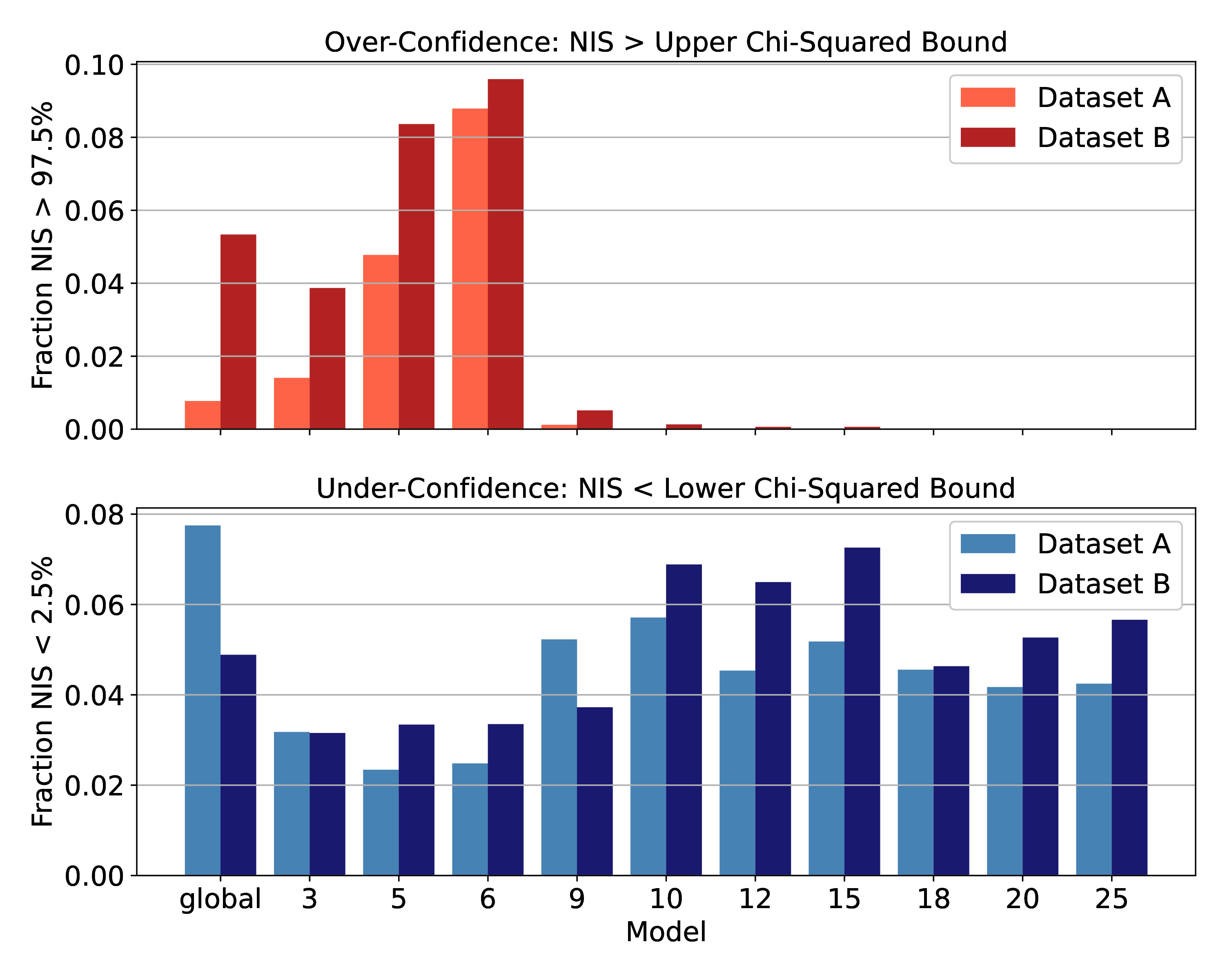}
    \caption{Bar plot of average number of samples per run that lie outside the confidence interval for the NIS statistics. The figure shows an initial deterioration in the filter performance when the number of components of the gaussian mixture are increased but later show improved NIS scores for higher number of components.}
    \label{fig:ModelWts_NIS}
\end{figure}

The NIS metrics validates filter consistency by calculating the value :
\begin{align}
    \nu_k &= \mathbf{y}_k^T \mathbf{S}_k^{-1}  \mathbf{y}_k \quad \& \quad 
    \nu = \frac{1}{L} \sum_{k=1}^{L} \nu_{k}
\end{align}
Where, at every step $k$ for the sequence length $L$, $\mathbf{\bar{y}}$ is know as \textit{innovation} or measurement residual, and, $\mathbf{S}$ is the innovation covariance which captures the total uncertainty (combination of state prediction and measurement uncertainty) in the state estimation. $\nu$ follows a chi squared distribution and the filter is validated for consistency with the $5\%$ and $95\%$ confidence interval. Fig.~\ref{fig:ModelWts_NIS} illustrates the results of utilizing different component gaussian mixtures during a state estimation run.

The figure illustrates the NIS statistics performance as the number of components of the gaussian mixture increases. Along side the NIS metrics, the figure also illustrates how the model weights fluctuate indicating utility of different models for the state estimation process. The figure illustrates the NIS statistics performance is subpar as compared to the single global model when the number of gaussian components are less than $9$. While this figure illustrates the phenomenon for one particular run, it can be useful to investigate the trend across all runs. 

Figure~\ref{fig:ModelWts_NIS} illustrates the NIS statistics for the global model in comparison to the n-component gaussian mixtures. The bars indicated the average number of samples that lie outside the confidence interval. The lighter shades of the bar indicate the validation on the dataset used for model fitting as described in section~\ref{sec:Dataset}. The darker shades on the other hand is validation on the dataset not used for tuning. Typically, for all runs the statistics on the seen dataset are better than the unseen dataset indicating the value of having accurate motion models. The upper model captures the statistics when the NIS values greater than the $97.5\%$ bound of the confidence interval (filter is overconfident) and the lower figure indicates the  NIS values  lower than the $2.5\%$ bound of the confidence interval (filter is underconfident.)

Overall, the filter is more underconfident than overconfident, which means the assumed noise values could be much higher which entails it would take noticeable time for the filter to converge. When aggregated for all runs, the figure indicates that NIS statistics worsen as compared to the single global model for up to $6$ component gaussian. For nine and above, the statistics are much better with components $10,12,15,18$ showing perfect scores.

Poor NIS statistics are typically driven due to model uncertainty or inaccurate measurement noise calibration. Since the measurement noise is kept same across all the runs, it is clear that the improvement in NIS scores is brought about by the improved prediction model accuracy or reduction in the motion model uncertainty due to the mixture of gaussian. This is a clear indicative that the GMM based modeling approach can be useful for identifying motion models for state estimation.

\section{Discussion}~\label{sec:Discussion}
In this work, a gaussian mixture model approach for model identification is blended with the interactive multiple model estimation framework for state estimation for a skid-steered wheel mobile robot. The proposed approach was presented as an alternative to the standard linear model with kalman filter which typically falls short for highly non-linear systems such as the skid-steering robots. The preliminary results indicate that the framework clearly performs much better as compared to the single model-kalman filter approach, at least from the point of view of measurement statistics.

While better in performance, one draw back of this approach can be the added efforts in identifying the several locally linear models through the entire dataset. The choice of the window size (chosen to be $25$ in this work) for identifying the locally linear models can be an additional tuning parameter which may dictate the performance. For future, a single step transition model (window size $1$) is proposed to be investigated to investigate the framework.

The choice of the number of components (chosen to be between $3$ to $25$) is yet another parameter that needs to be systematically investigated for future work. More critically, a covariance analysis which captures the uncertainty along with the means of the gaussian models also needs to be investigated for the state estimation performance.

Finally, the framework can benefit from investigating the state estimation performance in comparison to the ground truth. In that context, having improved sensors (such as GPS), and better instrumentation (high frequency noise free data sampling) can benefit for this investigation. A future work involving accurate RTK-GPS with a full scale tracked vehicle is proposed for validating the GMM-IMM framework.

\begin{ack}
Clemson University, Department of Mechanical and Automotive engineering acknowledge the technical and financial support of the Automotive Research Center (ARC) in accordance with Cooperative Agreement W56HZV-19-2-0001 US Army CCDC Ground Vehicle Systems Center (GVSC) Warren, MI.
\end{ack}

\bibliography{IMMIce}             
                                                   
\end{document}